\documentclass[submission,copyright,creativecommons]{eptcs}
\providecommand{\event}{ICLP DC 2019} 
\usepackage{breakurl}             
\usepackage{underscore}           
\usepackage{graphicx}
\usepackage{url}
\usepackage{color}
\usepackage{enumitem}
\usepackage{listings}
\usepackage{graphicx}
\usepackage[noend,ruled,linesnumbered]{algorithm2e} 

\title{Knowledge Authoring and Question Answering with KALM}
\author{Tiantian Gao
\institute{Department of Computer Science\\
Stony Brook University\\
Stony Brook, USA}
\email{tiagao@cs.stonybrook.edu}
}
\def\titlerunning{KALM for Knowledge Authoring and Question Answering}
\def\authorrunning{Tiantian Gao}
\begin{document}
\maketitle

\begin{abstract} 
Knowledge representation and reasoning (KRR) is one of the key areas in artificial intelligence (AI) field.
It is intended to represent the world knowledge in formal languages (e.g., Prolog, SPARQL) and then
enhance the expert systems to perform querying and inference tasks.
Currently, constructing large scale knowledge bases (KBs) with high quality is prohibited by the fact that
the construction process requires many qualified knowledge engineers who not only
understand the domain-specific knowledge
but also have sufficient skills in knowledge representation.
Unfortunately, qualified knowledge engineers are in short supply.
Therefore, it would be very useful to build a tool that allows the user to construct and query the KB simply via text. 
Although there is a number of systems developed for knowledge extraction and question answering,
they mainly fail in that these system don't achieve high enough accuracy whereas KRR is highly sensitive to erroneous data.
In this thesis proposal, I will present Knowledge Authoring Logic Machine (KALM), a rule-based system which
allows the user to author knowledge and query the KB in text. 
The experimental results show that KALM achieved superior accuracy in knowledge authoring and question answering
as compared to the state-of-the-art systems.
\end{abstract}

\section{Introduction}\label{intro}
Knowledge representation and reasoning (KRR) is the process of representing the domain knowledge in formal languages
(e.g., SPARQL, Prolog) such that it can be used by expert systems to execute
querying and reasoning services. 
KRR have been applied in many fields including financial regulations, medical diagnosis, laws, and so on.
One major obstacle in KRR is the creation of large-scale knowledge bases with high quality. 
For one thing, this requires the knowledge engineers (KEs) not only to have the background knowledge in a certain domain
but have enough skills in knowledge representation as well. Unfortunately, qualified KEs are
also in short supply.
Therefore, it would be useful to build a tool that allows the domain experts without any background in logic to construct and query the knowledge base
simply from text.

Controlled natural languages (CNLs) \cite{kuhn2014} were developed as a technology that achieves this goal. 
CNLs are designed based on natural languages (NLs) but with restricted syntax
and interpretation rules that determine the unique meaning of the sentence.
Representative CNLs include Attempto Controlled English \cite{fuchs2008} and PENG \cite{schwitter2010}.
Each CNL is developed with a language parser which translates the English sentences
into an intermediate structure, \emph{discourse representation structure} (DRS) \cite{kamp2013}.
Based on the DRS structure, the language parsers further translate the DRS into
the corresponding logical representations, e.g., \emph{Answer Set Programming} (\emph{ASP}) \cite{gebser2011} programs. 
One main issue with the aforementioned CNLs is that the systems do not provide
enough background knowledge to preserve semantic equivalences of sentences
that represent the same meaning but are expressed via different linguistic structures. For instance, 
the sentences \emph{Mary buys a car} and \emph{Mary makes a purchase of a car}
are translated into different logical representations by the current CNL parsers.
As a result, if the user ask a question \emph{who is a buyer of a car}, these systems
will fail to find the answer.

In this thesis proposal, I will present KALM \cite{Gao18,GaoFK18}, a system for knowledge authoring and question answering.
KALM is superior to the current CNL systems in that KALM has a complex frame-semantic parser
which can standardize the semantics of the sentences that express the same meaning
via different linguistic structures. The frame-semantic parser is built based on FrameNet \cite{JohnsonEtAl:01a}
and BabelNet \cite{NavigliPonzetto:12aij} where 
FrameNet is used to capture the meaning of the sentence and BabelNet \cite{NavigliPonzetto:12aij}
is used to disambiguate the meaning of the extracted entities from the sentence.
Experiment results show that KALM achieves superior accuracy in knowledge authoring and question answering
as compared to the state-of-the-art systems.

The rest parts are organized as follows: 
Section \ref{kalm} presents the KALM architecture,
Section \ref{kalm-qa} presents KALM-QA, the question answering part of KALM,
Section \ref{experiment} shows the evaluation results,
Section \ref{related} discusses the related works, 
Section \ref{future} shows the future work beyond the thesis,
and Section \ref{conclusion} concludes the paper.

\section{The KALM Architecture}\label{kalm}
\begin{figure}[h]
 \centering
 \includegraphics[width=11cm]{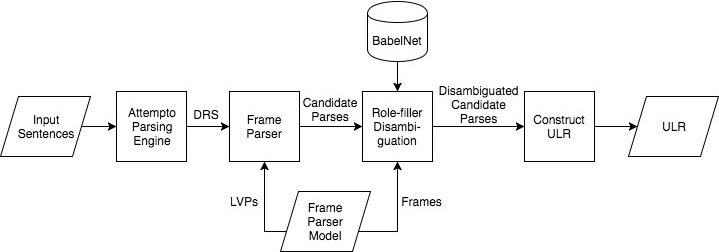}
 \caption{The KALM Architecture}
 \label{sentencetoulr}
\end{figure}

Figure \ref{sentencetoulr} shows the architecture of KALM which translates a CNL sentence to the 
corresponding logical representations,
\emph{unique logical representations} (\emph{ULR}).

\noindent
\textbf{Attempto Parsing Engine.} The input sentences are CNL sentences based on ACE grammar.\footnote{\url{http://attempto.ifi.uzh.ch/site/docs/syntax_report.html}}
KALM starts with parsing 
the input sentence using ACE Parser\footnote{\url{https://github.com/Attempto/APE}}
 and generates the DRS structure \cite{fuchs2010} which captures the syntactic information of the sentences.

\noindent
\textbf{Frame Parser.} KALM performs
frame-based parsing based on the DRS and produces a set of frames that represent the semantic relations 
a sentence implies. A frame \cite{fillmore01:FStextUnder} represents a semantic relation
of a set of entities where each plays a particular role in the frame relation.
We have designed a frame ontology, called \emph{FrameOnt}, which is based on the frames
in FrameNet \cite{JohnsonEtAl:01a} and encoded as a Prolog fact. 
For instance, the \texttt{Commerce\_Buy} frame is shown below:
\begin{verbatim}
    fp(Commerce_Buy,[
       role(Buyer,[bn:00014332n],[]),
       role(Seller,[bn:00053479n],[]),
       role(Goods,[bn:00006126n,bn:00021045n],[]),
       role(Recipient,[bn:00066495n],[]),
       role(Money,[bn:00017803n],[currency])]).
\end{verbatim}
In each \texttt{role}-term, 
the first argument is the name of the role and 
the second is a list of \emph{role meanings} represented via
BabelNet synset IDs \cite{NavigliPonzetto:12aij}.
The third argument of a \texttt{role}-term
is a list of constraints on that role.
For instance, the sentence \emph{Mary buys a car} implies the \texttt{Commerce\_Buy} frame where \emph{Mary}
is the \texttt{Buyer} and \emph{car} is the \texttt{Goods}.
To extract a frame instance from a given CNL sentence, KALM uses \emph{logical valence patterns} (\emph{lvps}) which 
are learned via structural learning. An example of the lvp is shown below:
\begin{verbatim}
    lvp(buy,v,Commerce_Buy, [
      pattern(Buyer,verb->subject,required),
      pattern(Goods,verb->object,required),
      pattern(Recipient,verb->pp[for]->dep,optnl),
      pattern(Money,verb->pp[for]->dep,optnl),
      pattern(Seller,verb->pp[from]->dep,optnl)]).
\end{verbatim}
\noindent
The first three arguments of an \texttt{lvp}-fact identify the
lexical unit, its part of speech, and the frame.
The fourth argument is a set of \texttt{pattern}-terms,
each having three parts: 
the name of a role, 
a grammatical pattern, and the required/optional flag.
The \emph{grammatical pattern} 
determines the grammatical
context in which
the lexical unit, a role, and a role-filler word can appear in that frame.
Each grammatical pattern is captured by a \emph{parsing rule} (a Prolog
rule) that  
can be used to extract appropriate role-filler words based on the APE parses.

\noindent
\textbf{Role-filler Disambiguation.} Based on the extracted frame instance, the role-filler disambiguation module
disambiguates the meaning of each role-filler word for the corresponding frame role a BabelNet Synset ID.
A complex algorithm \cite{Gao18} was proposed to measure the semantic similarity between a candidate BabelNet synset
that contains the role-filler word and the frame-role synset. The algorithm also has optimizations that improve
the efficiency of the algorithm e.g., priority-based search, caching, and so on.
In addition to disambiguating the meaning of the role-fillers, this module is also used to prune the extracted
frame instances where the role-filler word and the frame role are semantically incompatible.

\noindent
\textbf{Constructing ULR.} The extracted frame instances are translated into the corresponding logical representations,
\emph{unique logical representation} (\emph{ULR}). Examples can be found in reference  \cite{Gao18}.

\section{KALM-QA for Question Answering}\label{kalm-qa}
\begin{figure}[h]
 \centering
 \includegraphics[width=14cm]{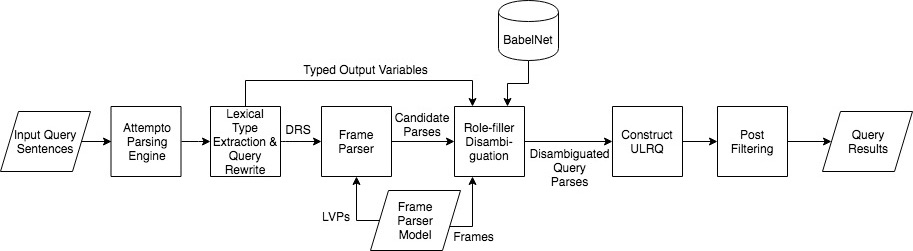}
 \caption{The KALM-QA Architecture}
 \label{kalmqa}
\end{figure}

Based on KALM, KALM-QA \cite{GaoFK18} is developed for question answering. Figure \ref{kalmqa} shows the KALM-QA
architecture.
KALM-QA shares  the same components
with KALM for syntactic parsing, frame-based parsing and role-filler disambiguation.
Different from KALM, KALM-QA translates the questions to \emph{unique logical representation for queries}
 (\emph{ULRQ}), which are used to query the authored knowledge base.

\section{Evaluations}\label{experiment}
This section provides a summary of the evaluation of KALM and KALM-QA, where KALM is evaluated for knowledge authoring
and KALM-QA is evaluated for question answering.
We have created a total of 50 logical frames, mostly derived from FrameNet but also
some that FrameNet is missing (like Restaurant, Human\_Gender) for representing the meaning of 
English sentences. Based on the 50 frames, we have manually constructed 250 sentences
that are adapted from FrameNet exemplar sentences and evaluate these sentences on KALM, 
SEMAFOR, SLING, and Stanford KBP system. 

The evaluation is based on the following metrics:

\vspace{1mm}
\noindent
\begin{tabular}{ p{2.3cm} p{13cm}}
\quad FrSynC  &  all frames and roles (semantic relations) are
identified correctly and all role-fillers are disambiguated\\
\quad FrC  & all frames and roles are identified correctly   \\
\quad PFrC & some frames/roles are identified, but some are not \\
\quad Wrong & some frames or roles are misidentified\\
\end{tabular}

\vspace{0.5cm}
The results are shown as follow:
\begin{description}
\item[\quad \it KALM:] 239 sentences are \emph{FrSynC} (\textbf{95.6}\%),
248 sentences are \emph{FrC} ($>$ \textbf{99}\%), and
2 sentences are \emph{Wrong} ($<$ 1\%).
Note that \emph{FrSynC} applies only to KALM,
since none of the comparison systems can disambiguate
the senses of the extracted entities.
\item[\quad \it SEMAFOR:] parses 236 sentences out of the 250  test sentences,
where 59 sentences are \emph{FrC} (\textbf{25}\%),
44 sentences are \emph{PFrC} (18.6\%),
and 133 sentences are \emph{Wrong} (56.4\%).
\item[\quad \it SLING:] parses 233 sentences, where
98 sentences are \emph{FrC} (\textbf{42.1}\%),
63 are \emph{PFrC} (27\%),
and 72 sentences are \emph{Wrong} (30.9\%).
\item[\quad \it Stanford KBP:] parses 26 sentences, where
14 sentences are \emph{FrC} (\textbf{53.8}\%),
10 sentences are \emph{PrC} (38.5\%),
and 2 sentences are \emph{Wrong} (7.7\%).
\end{description}

The differences between KALM and other systems are listed in order.
First, none of the other systems 
do disambiguation or attempt to find synsets for role-fillers, so in
this aspect KALM does more and is better attuned to the task of knowledge
authoring.
Second, none of these systems can explain their results,
nor do they provide ways to analyze and correct errors.
Third, KALM achieves an accuracy of 95.6\%---much higher than other systems.

For KALM-QA, we evaluate it on two datasets. The first dataset is manually constructed general questions
based on the 50 logical frames. KALM-QA achieves an accuracy of 95\% for parsing the queries.
The second dataset we use is MetaQA dataset \cite{ZhangDKSS18}, which contains 
contains almost 29,000 test questions and
over 260,000 training questions. KALM-QA achieves 100\% accuracy---much higher than the state-of-the-art
machine learning approach \cite{ZhangDKSS18}. Details of the evaluations can be found in \cite{Gao18} and \cite{GaoFK18}.

\section{Related Works}\label{related}
As is described in Section \ref{intro}, CNL systems were proposed as the technology for 
knowledge representation and reasoning.
Related works also include knowledge extraction tools, e.g., OpenIE \cite{angeli2015}, SEMAFOR \cite{journals/coling/DasCMSS14}, 
SLING \cite{ringgaard2017sling}, and Standford KBP system \cite{ManningSBFBM14}.
These knowledge extraction tools are designed to extract semantic relations from English sentences that capture the 
meaning. The limitations of these tools are two-fold:
first, they lack sufficient accuracy to extract the correct semantic relations and entities
while KRR is very sensitive to incorrect data;
second, these systems are not able to map the semantic relations to logical forms and therefore
not capable of doing KRR.
Other related works include the question answering frameworks, e.g., Memory Network \cite{MillerFDKBW16}, Variational Reasoning Network \cite{ZhangDKSS18}, 
ATHENA \cite{SahaFSMMO16}, PowerAqua \cite{LopezFMS12}. 
The first two belong to end-to-end learning approaches based on machine learning 
models. The last two systems have implemented semantic parsers which translate natural language sentences
into intermediate query languages and then query the knowledge base to get the answers.
For the machine learning based approaches, the results are not explainable. Besides,
their accuracy is not high enough to provide correct answers.
For ATHENA and PowerAqua, these systems perform question answering based on a priori knowledge bases. 
Therefore, they do not support knowledge authoring while KALM is able to support both knowledge authoring and question answering.

\section{Future Work Beyond The Thesis}\label{future}
This section discusses the future work beyond the thesis:  (1) enhancing KALM to author rules, 
and (2) supporting time reasoning.

\noindent
\textbf{Authoring Rules from CNL.} There are two research problems with rules.
The first problem is the standardization of rules parses that express the same
information but via different syntactic forms or using different expressions.
Suppose the knowledge base contains sentences like:
(1) \emph{if a person buys a car then the person owns the car},
(2) \emph{every person who is a purchaser of a car is an owner of the car},
(3) \emph{if a car is bought by a person then the person possesses the car}.
All the above sentences represent rules and express exactly the same meaning.
However, KALM's current syntactic parser will represent them in different DRSs and therefore not being
able to map them into the same logical form.
The second problem involves the recognition and representation of different types of rules in logic.
For instance, defeasible rules are very common in text. However, this type of rules
cannot be handled by first order logic. We believe defeasible logic \cite{wan2009} is a good fit.

\noindent
\textbf{Time Reasoning.} Time-related information is a crucial part of human
knowledge, but semantic parsing that takes the time into account is rather
hard. However, we can develop a CNL that would incorporate enough time related
idioms to be useful in a number of domains of discourse (e.g., tax
law). Time can then be added to DRSs and incorporated into our frame based
approach down to the very level of the logical facts into which sentences
will be translated. This time information can be represented either via special
time-aware relations among events (e.g., before, after, causality, triggering)
or using a reserved argument to represent time in each fluent.

\section{Conclusions}\label{conclusion}
This thesis proposal provides an overview of KALM, a system for knowledge authoring. In addition,
it introduces KALM-QA, the question answering part of KALM. Experimental results show
that both KALM and KALM-QA achieve superior accuracy as compared to the state-of-the-art systems.

\bibliographystyle{eptcs}
\bibliography{main}
\end{document}